\documentclass[sigconf]{acmart}
\AtBeginDocument{%
  }

\setcopyright{none}
\renewcommand\footnotetextcopyrightpermission[1]{}
\acmConference[C+J '24]{Computation + Journalism 2024}{October 25-27, 2024}{Boston, MA}
\usepackage{lipsum}
\usepackage{tablefootnote}
\usepackage{multirow}
\usepackage{array}
\usepackage{tabularx}
\usepackage{enumitem}
\usepackage{xcolor,colortbl}
\usepackage{fontawesome}

\begin{document}

\definecolor{analyst}{HTML}{d3e1f1} 
\definecolor{reporter}{HTML}{ffdbac}
\definecolor{editor}{HTML}{efd6d6} 
\definecolor{sota}{HTML}{efd6d6} 

\title{Using Generative Agents to Create Tip Sheets for Investigative Data Reporting}

\author{Joris Veerbeek}
\email{j.veerbeek@uu.nl}
\affiliation{%
  \institution{Utrecht University}
   \city{Utrecht}
  \country{The Netherlands}
}

\author{Nicholas Diakopoulos}
\email{nad@northwestern.edu}
\affiliation{%
  \institution{Northwestern University}
     \city{Evanston}
  \country{USA}
}

\begin{abstract}
 This paper introduces a system using generative AI agents to create tip sheets for investigative data reporting. Our system employs three specialized agents—an analyst, a reporter, and an editor—to collaboratively generate and refine tips from datasets. We validate this approach using real-world investigative stories, demonstrating that our agent-based system generally generates more newsworthy and accurate insights compared to a baseline model without agents, although some variability was noted between different stories. Our findings highlight the potential of generative AI to provide leads for investigative data reporting.  

\end{abstract}


\maketitle
\section{Introduction}

To what extent can generative AI models uncover noteworthy insights from datasets that might offer leads for investigative data reporting? Long before generative AI gained widespread attention, computational journalism envisioned systems that would alert reporters to trends and anomalies in vast streams of data \cite{hamilton2009accountability}. Over the past decade, this vision materialized through various prototypes \cite{broussard2015artificial}, from identifying fact-checkable statements in transcripts to automatically monitoring data streams using statistical methods \cite{diakopoulos2020computational}. Commonly referred to as \textit{computational news discovery} (CND), these approaches are defined as ``the use of algorithms to
orient editorial attention to potentially newsworthy events
or information prior to publication'' \cite{diakopoulos2020computational}. Now, large language models (LLMs) hold the potential not only to identify more complex newsworthy patterns in datasets but also to \textit{generate} news angles with greater flexibility and creativity, overcoming the limitations of standard templates \cite{nishal2024understanding}.

In this work, we design, develop, and evaluate a system employing LLMs to generate ``tip sheets'' that provide a list of noteworthy observations that may inspire further journalistic exploration of datasets \citep{diakopoulos2020generating}. This system uses OpenAI’s Assistants API, which allows GPT-4 to execute code and interact with the results of data analyses \cite{openai2024}. GPT-4's capabilities in analyzing and interpreting datasets have been evaluated in various contexts, including applied data science \cite{cheng2023gpt, rasheed2024can}, visual exploratory data analysis \cite{stigall2023chatbots}, and qualitative data analysis \cite{rasheed2024can}. Unlike those evaluations, the pipeline we present in this paper is specifically designed to facilitate investigative data journalism tasks and uncover newsworthy insights within datasets.

We aim to align the analytical capabilities of LLMs with the objectives of investigative data journalism by modeling them as a set of generative agents \cite{park2023generative}. Similar to recent work on generative literary translation \cite{rasheed2024can} and text interpretation \cite{rasheed2024can}, we developed AI agents with specialized roles designed to perform distinct subtasks and offer mutual feedback. In our system, these roles resemble those of a data analyst, an investigative reporter, and a data editor. We validate our generative agents pipeline against real-world investigative data reporting stories and compare the outcomes to a baseline setup that lacks such agents. Our evaluation demonstrates that using generative agents significantly enhances newsworthiness and improves the validity of findings.

\begin{figure}[t!]
  \centering
  \includegraphics[width=0.6\linewidth]{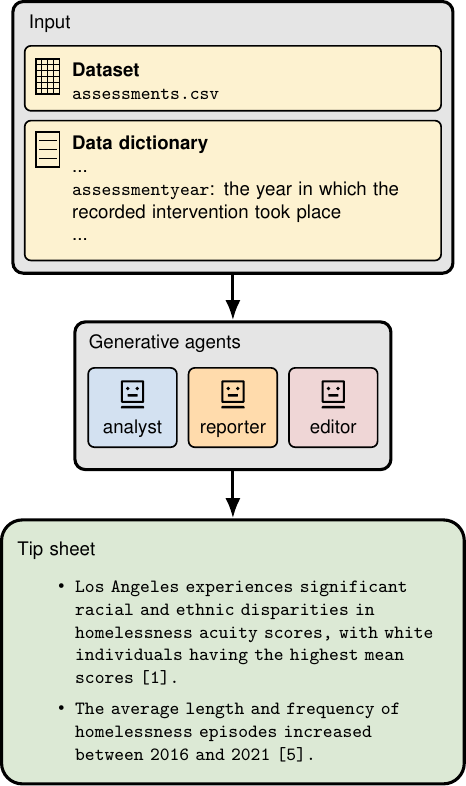}
  \vspace{0em}
  \caption{Given a dataset and a description of the dataset, our generative agents pipeline outputs a tip sheet.}
  \label{fig:teaser}
\end{figure}

\section{Generative Agents Pipeline}

Our system employs three distinct agents: an analyst, a reporter, and an editor, each fulfilling specific tasks as described in Table \ref{tab:roles}. The analyst performs data analysis, the reporter generates questions and summarizes findings, and the editor ensures integrity and verifies the work. Both the reporter and analyst have access to data, while the editor is equipped with document retrieval functions, which allow it to access three general sets of guidelines for bulletproofing data-driven work. Overall, the user provides a dataset and a data description, and the final output is a tip sheet of potentially newsworthy insights from the data that could be pursued by the user. The intermediate process is divided into four key steps:

\begin{enumerate}
\item \textbf{Question Generation}: The reporter agent generates a set of questions that are feasible to address using the provided dataset.
\item \textbf{Analytical Planning}: For each question, the analyst drafts an analytical plan detailing how the dataset can be used to answer the question.
\item \textbf{Execution and Interpretation}: Each analytical plan is executed and interpreted by the analyst. The editor and reporter provide feedback, which the analyst incorporates, and the reporter then summarizes the final results in bullet points.
\item \textbf{Compilation and Presentation}: All bullet points are compiled, and a subset of the most significant findings is presented to the user in the tip sheet. 
\end{enumerate}


The development of these steps and the corresponding prompts involved iterative refinement and strategic approaches. For example, we looked at job adverts to help write the system prompts. Other aspects, such as avoiding interview suggestions, producing visualizations, or limiting feedback suggestions, were established through trial and error on a data journalism story outside our test set. Nonetheless, we recognize that the pipeline presented here is just one possible approach, not necessarily the only or most optimal one.

In the next sections, we provide a detailed description of each step. A comprehensive overview of the process is shown in Figure \ref{fig:diagram_large}, which we reference throughout the remainder of this section.\footnote{A complete overview, including the prompts used by our generative agent-based system and the associated code, is available at \texttt{https://github.com/veerbeek/agents} }

\subsection{Question Generation (Step 1)}
To initiate a project, the user must prepare two files: (1) a dataset in CSV format, and (2) a Markdown file containing a general description of the dataset and a data dictionary that includes comprehensive explanations of each variable in the dataset.

\begin{figure*}[t] %
  \centering
\includegraphics[width=0.85\textwidth,height=0.75\textheight,keepaspectratio]{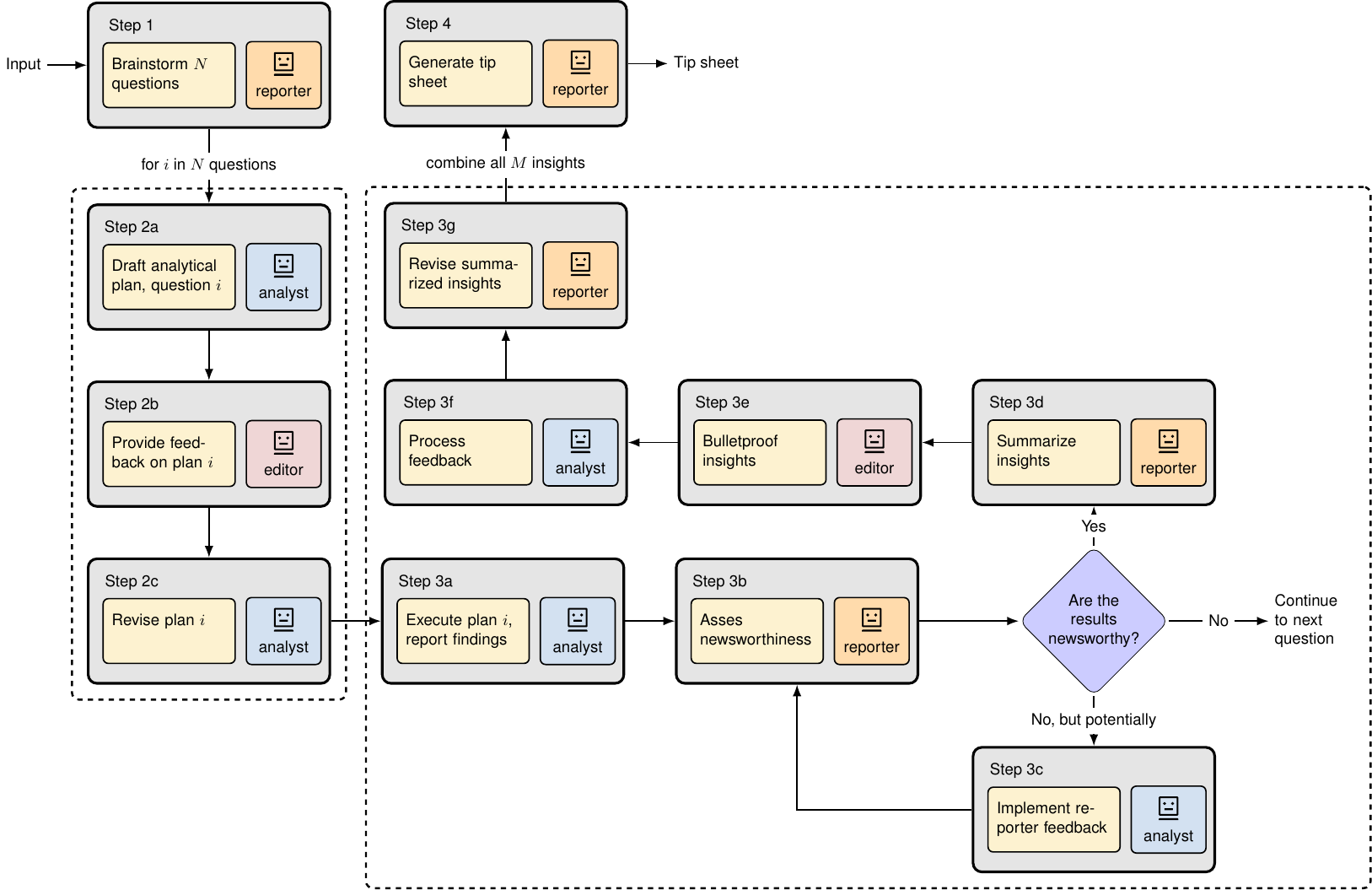}
  \caption{A complete overview of the pipeline. Each box represents a single prompt assigned to one of the three agents.}
  \label{fig:diagram_large}
\end{figure*}

Additionally, users can configure five parameters: the number of questions generated (default: 10), the number of bullet points in the final tip sheet (default: 10), the maximum number of interactions between analyst and reporter agents (default: 3), and whether to utilize the editor and reporter agents (both default to true). 

To initialize the agent pipeline, the dataset description is provided to the reporter agent, accompanied by a prompt to: (1) explore the dataset by shuffling it, printing the columns, and viewing the head of the dataframe, and (2) generate $N$ newsworthy questions. The reporter is instructed to number the questions, which are then used to parse the agent's response and organize the questions into a list.

\begingroup

\begin{table}
  \caption{Roles and tasks of the different agents in our pipeline}
  \label{tab:roles}
  \centering
  \begin{tabular}{p{0.8cm} >{\raggedleft\arraybackslash}p{0.6cm} >{\raggedleft\arraybackslash}p{0.9cm} p{4.5cm}}
    \toprule
    Agent & Data & Retrieval & General tasks \\
    \midrule
    \texttt{analyst} &  \includegraphics[scale=0.25]{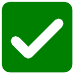} & \includegraphics[scale=0.17]{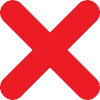}  & \vspace{-1em}
    \begin{itemize}[leftmargin=*, topsep=0pt]
      \setlength\itemsep{0em}
        \item Translating journalistic questions into quantitative analyses.
        \item Conducting the data analysis.
        \item Interpreting analysis results.
    \end{itemize} \\
    \texttt{reporter} & \includegraphics[scale=0.25]{assets/check_mark.pdf} & \includegraphics[scale=0.17]{assets/red_cross.pdf} & \vspace{-1em}
    \begin{itemize}[leftmargin=*, topsep=-1pt, partopsep=0pt]
        \item Generating relevant and newsworthy questions.
        \item Challenging the data analyst agent with follow-up questions.
        \item Summarize the insights.
    \end{itemize} \\
    \texttt{editor} & \includegraphics[scale=0.17]{assets/red_cross.pdf} & \includegraphics[scale=0.25]{assets/check_mark.pdf} & \vspace{-1em}
    \begin{itemize}[leftmargin=*, topsep=0pt, partopsep=0pt]
        \item Guarding the journalistic integrity of the process.
        \item Bulletproofing analytical plan.
        \item Checking findings of analyst and reporter for factual accuracy.
    \end{itemize} \\
    \bottomrule
  \end{tabular}
\end{table}

\subsection{Analytical Planning (Step 2)}
After the questions are brainstormed, the analyst is next to take action. For each question, the analyst is provided with the dataset description and is prompted to draft an analytical plan. This plan includes the entire data process from start to finish, such as preparing, cleaning, and checking the data. Once this plan is prepared, it is forwarded to the editor agent, who is instructed to review the plan critically, drawing upon its resources about bulletproofing data journalism stories in its knowledge base. The feedback from the editor agent is then sent back to the analyst, who is asked to revise the plan based on this input. This revised plan becomes the final analytical plan to be used in subsequent stages of the process.

\subsection{Execution and Interpretation (Step 3)}
Once the analytical plan is finalized, it is sent to the analyst agent with instructions to execute the plan. After the plan is executed, the analyst is instructed to summarize its approach and the resulting insights into bullet points. This additional step is necessary because the output of the execution often spans multiple messages and incorporates extensive Python code.

\subsubsection{Feedback stage 1: Reporter feedback (Step 3b)}
The bullet points summarizing the approach and the analysis are then forwarded to the reporter agent. When receiving an analysis, the reporter is instructed to assess it under one of three possible options:

\begin{enumerate}
\item The analysis yields newsworthy insights for publication.
\item The analysis does not currently offer enough newsworthy insights, but merits additional investigation to explore other potential angles.
\item The analysis lacks newsworthy insights and is unlikely to produce any from the dataset given this question.
\end{enumerate}

The reporter's response always includes the chosen option and, if necessary, specific feedback on the analysis along with suggestions for additional angles to investigate. If option 2 is selected by the reporter, the analyst is prompted to conduct a new analysis incorporating the feedback provided by the reporter (Step 3c). These results are again summarized into bullet points and sent to the reporter for feedback.   

When option 3 is chosen, or the maximum number of feedback loops is reached, the agents cease working on the current question and move on to the next one. Only if option 1 is selected do we proceed to the next stage of feedback (Step 3d).

\subsubsection{Feedback stage 2: Editor feedback (Step 3e)}
 After option 1 is chosen, the bullet points are forwarded to the editor. The editor reviews these points and provides feedback on the approach and analysis, possibly requesting additional checks and analyses. The analyst receives this feedback (Step 3f), implements the necessary changes, and then again summarizes their revised approach and insights in bullet points.

\newcolumntype{b}{X}
\newcolumntype{m}{>{\hsize=.4\hsize}X}
\newcolumntype{s}{>{\hsize=.4\hsize}X}

\begin{table*}[t]
  \caption{The five projects included in the evaluation our generative agents pipeline}
  \label{tab:projects}
  \begin{tabularx}{\textwidth}{m s b s}
    \toprule
    Project&Organization&Topic&Data category\\
    \midrule
    \texttt{themarkup-scoring}\footnotemark& The Markup (USA) & LA’s scoring system for people experiencing homelessness
 & Survey data  \\ 
    \texttt{rferl-headlines}\footnotemark & RFE/RL (EU) & Media headlines from Hungary and Russia
    & Traditional media \\ 
    \texttt{civio-emergency}\footnotemark & Civio (SP) & Emergency contracts during the pandemic
    & Governmental \\
    \texttt{netra-executions}\footnotemark & Netra (SE/BD) & Extrajudicial executions Bangladesh
    & Judicial \\ 
    \texttt{readr-ads}\footnotemark & Readr (TW) & Political ads on Facebook
    & Social media  \\ [0.1cm]

  \bottomrule
\end{tabularx}
\end{table*}

Finally, all the bullet points generated by the analyst throughout the entire process of answering a question are sent to the reporter. The reporter is prompted to summarize the most newsworthy insights into bullet points. These insights are stored for generating the tip sheet later, and the process starting from step 2 begins anew with the next question.

\subsection{Compilation and Presentation (Step 4)}
After completing steps 2 and 3 for all questions, each question yields a list of bullet points that capture its most significant findings. In the final phase of the process, these bullet points are numbered for each question and given to the reporter. The reporter is then responsible for summarizing the most newsworthy insights from all the analyses into a predetermined number of bullet points. 
This curated list of bullet points constitutes the tip sheet.

\section{Evaluation}
Drawing on recent work on the importance of evaluating LLMs in a domain-specific context rather than general benchmarks \cite{nishal2024domain}, we assess the quality of the tip sheet by manually gathering datasets of real-world investigative data reporting. Using these datasets, we generate tips and evaluate their validity and newsworthiness. Additionally, we compare the tips generated by LLMs to the findings presented in the actual stories.

\subsection{Dataset selection}
Table \ref{tab:projects} provides an overview of the projects that we use to evaluate our setup. For the selection of the projects, we aimed for diversity in publication location, methods, and types of insights. All projects were nominated for the Sigma Awards or the Philip Meyer Journalism Award and thus reflect high-quality data journalism. We include projects that featured multiple clear data-driven conclusions rather than general mappings of phenomena, with data that was publicly available or made accessible. Additionally, the key dataset for each project had to fit within a single CSV file. Given the limitations set by OpenAI's Assistants API, projects were excluded if they required intensive computational resources, used very large datasets, involved visual image analysis, or focused primarily on geographical analysis. 

\subsection{Evaluation metrics}
The advantage of using real-world projects is that we can compare our findings with the data-driven claims made in the published articles. By comparing the claims made by our agents to those featured in the story, we achieve a measure similar to \textit{precision} in information retrieval. To determine if a generated claim is equivalent to a claim in the article, we do not require the wording or even methodology to be identical. Instead, alignment depends on whether the described insight is similar based on relationships between variables, categorical distinctions, rankings, or the mention of specific numerical values, such as the total number of items in the dataset.

While precision is an important metric, the absence of a claim in an article does not necessarily imply that the claim is irrelevant or not potentially newsworthy. Without interviewing the reporters who worked on each project in depth we cannot know which claims were omitted or why; the story might have focused on certain aspects for a variety of editorial reasons \cite{Harcup:2016bn}, while our agents highlighted others. Therefore, we also assess the \textit{newsworthiness} of claims not included in the article. Since newsworthiness can be challenging to uniformly define across various contexts and institutions, we only determine if a claim could be considered \textit{potentially} newsworthy or not. This binary rating is thus indicative rather than suggesting that a finding would necessarily result in a news item. We apply the news values described by \cite{diakopoulos2021towards} to determine potential newsworthiness. A claim is deemed potentially newsworthy if it aligns with one or more of the following news values: timeliness, power elite, relevance, bad news, magnitude, controversy, surprise, and actuality.

Finally, we assess the \textit{validity} of the claims produced by our agents. A claim is considered valid when it can be viewed as a reasonable inference based on the provided data. In practice, we assess this by manually checking the dataset to see if we can identify the same insights as described in each tip generated.

\addtocounter{footnote}{-5} 
\stepcounter{footnote}\footnotetext{https://themarkup.org/investigation/2023/02/28/l-a-s-scoring-system-for-subsidized-housing-gives-black-and-latino-people-experiencing-homelessness-lower-priority-scores}
\stepcounter{footnote}\footnotetext{https://www.currenttime.tv/a/peacekeepers-putin-and-orban-ru-hu-media-analysis/32408840.html}
\stepcounter{footnote}\footnotetext{https://civio.es/quien-cobra-la-obra/2021/03/23/four-companies-won-one-in-ten-euros-from-2020-emergency-contracts/}
\stepcounter{footnote}\footnotetext{https://interactive.netra.news/extrajudicial-killings-bangladesh}
\stepcounter{footnote}\footnotetext{https://www.readr.tw/project/political-post}

\subsection{Baseline}
To assess the benefits of using multiple agents, we compared our generative agents pipeline to a baseline that lacks these agents. To ensure a fair comparison, we designed the baseline model to answer the same number of questions as the agents in the pipeline and output a similar type of tip sheet. Specifically, this means that steps 1 and 4 in Figure \ref{fig:diagram_large} remain the same, but steps 2 and 3 are replaced by a single prompt asking the model to answer the question based on the dataset. The baseline model does not have a specific system prompt but receives exactly the same information about the dataset as the generative agents.

\subsection{Experiments}
We ran both the baseline and the generative agents setup using the \texttt{gpt-4-turbo-preview} model with a temperature setting of $1$. Due to the variability introduced by this temperature setting, we ran each story three times over the course of two weeks in mid-May 2024. All other parameters in the pipeline were set to their default values. In future work, we plan to evaluate the effects of these different parameters and various components of the pipeline. 

After running each setup, we code all tips according to our three evaluation metrics. The tips in the tip sheets are coded blindly and in random order, without regard to the setup. Only when a tip lacks sufficient information to assess its validity do we examine the associated code and interpretation.


\section{Results}
Table \ref{tab:results} shows the performance of both the baseline and the generative agents setup on our five selected data stories. In total, we evaluated 300 tips, with each story having 30 tips per setup. The results for each project are presented individually, with aggregated scores for all five projects provided at the end. Generally, the validity of the analyses produced by the models ranges between 0.80 and 0.90, which is relatively consistent with the percentages reported in comparable settings \cite{cheng2023gpt}. However, \texttt{rferl-headlines} is a significant outlier, likely due to its requirement for more complex text analysis methods in Hungarian and Russian. Despite this, the overall scores confirm the benefits of our agents pipeline, with higher aggregate scores across all three metrics. Particularly the difference in newsworthiness is notable, with consistently higher scores across all five projects. 

For the other two metrics, the results are less consistent across projects. Most notably, for \texttt{civio-emergency}, both the validity and precision are lower for the generative agents pipeline. As noted by the original authors and included in our dataset description, this dataset is relatively messy and might contain incorrect values. While this does not pose much of a problem when examining these values at an aggregate level, as done in the original article, it leads to incorrect interpretations when analyzing interactions between variables, which is often suggested by both the reporter and the editor as a possible next step. The possibility of data discrepancies was sometimes noted by the analyst in the analysis itself but was not included in the final tip sheet. For \texttt{themarkup-scoring}, we also observe a higher validity for the baseline, although the difference is relatively negligible considering the total number of tips included.

Additionally, for \texttt{readr-ads}, the baseline includes a higher percentage of findings from the original article compared to the agents setup. In this context, we observed that the agents quickly began to analyze the effectiveness (e.g., calculating the cost per impression) of the advertisements, shifting the focus more to the role of the platform. Although similar analyses have been conducted in various journalistic stories, the actual article primarily focused on the positions of the parties in Taiwan, which was better reflected in the baseline tips.

\begin{table}[t!]
\setlength\tabcolsep{5pt}

    \centering
    \caption{Evaluation of both setups on five selected stories. BL = baseline, GA = generative agents}
    \label{tab:results}
    \begin{tabular}{lccccccccccc}
        \toprule
        \multirow{2}{*}{Project} & \multicolumn{2}{c}{Validity} & \multicolumn{2}{c}{Newsw.} & \multicolumn{2}{c}{Precision} \\
        \cmidrule(r){2-3} \cmidrule(r){4-5} \cmidrule(r){6-7}
         & BL & GA & BL & GA & BL & GA \\
        \midrule
        \texttt{themarkup-scoring} & \cellcolor{sota}0.93 & 0.90 & 0.46 & \cellcolor{sota}0.70 & 0.04 & \cellcolor{sota}0.13 \\
        \texttt{rferl-headlines} & 0.57 & \cellcolor{sota}0.77 & 0.40 & \cellcolor{sota}0.63 & 0.33 & \cellcolor{sota}0.53 \\
        \texttt{civio-emergency} & \cellcolor{sota}1.00 & 0.93 & 0.63 & \cellcolor{sota}0.73 & \cellcolor{sota}0.40 & 0.27 \\
        \texttt{netra-executions} & 0.80 & \cellcolor{sota}0.87 & 0.47 & \cellcolor{sota}0.63 & 0.33 & \cellcolor{sota} 0.57 \\
        \texttt{readr-ads} & 0.80 & \cellcolor{sota}0.97 & 0.47 & \cellcolor{sota}0.67 & \cellcolor{sota}0.27 & 0.20 \\
        \midrule
        \textbf{Overall} & 0.82 & \cellcolor{sota} 0.89 & 0.52 & \cellcolor{sota}0.67 & 0.28 & \cellcolor{sota}0.34 \\
        \bottomrule
    \end{tabular}
\end{table}

\section{Discussion}
Although our evaluation highlighted the potential of generative agents for generating leads in investigative data reporting, a considerable portion of work in the field was not included in our assessment. The selection of projects, for example, was significantly influenced by the limitations of the OpenAI's assistant's API, which currently does not support executing complex code or using external packages that are not pre-installed in the environment. Consequently, many investigative data projects were excluded from consideration. In the future, non-proprietary models which can execute code in trusted sandboxes might be substituted to provide a greater range of options to the analyst agent. Additionally, data collection, another vital aspect of an investigative data reporter's work, was also omitted. The datasets provided to the agents in this study were relatively ready-to-use, whereas in many cases, considerable effort is required to obtain and clean these datasets. This evaluation solely focuses on the data analysis part, and future work could potentially assess the potential for agentic AI systems across the various stages of investigative data reporting \cite{showakat_baumer2021}.

Additionally, our evaluation mainly focused on what the models \textit{retrieved}, not on what they \textit{missed}. This includes the need for a more thorough typology of the types of insights the models found, such as trends, values, outliers, or other types as established in prior work \cite{Wang2019}. This would give an indication of the biases of the models towards certain findings, particularly when compared to the key findings highlighted in the stories. Future research could address these aspects to provide a more comprehensive and qualitative evaluation of the limitations of generative agents in providing leads for investigative data reporting.

In this exploratory evaluation, we pitted simple baseline models against the full pipeline. Future work could address the benefits of different components of the pipeline, such as the knowledge bases used, system prompts, and feedback loops, as well as varying different parameters. Finally, our current setup leaves limited agency for reporters. When applying this in a real newsroom, it would be essential to provide reporters with greater control via additional input possibilities. For example, this could be achieved by allowing them to provide input to or participate in the brainstorming phase.

\section{Conclusion}
In this work, we demonstrated the potential of using generative agents to create tip sheets for investigative data reporting. Our setup, comprising an analyst, reporter, and editor, was evaluated against a baseline without agents using real-world data stories. Although there was some variation between projects, the overall performance of our agents surpassed the baseline, particularly in terms of newsworthiness. 
At the same time, typically only a third of the findings generated end up somewhere in the final article. This underscores the role of the editorial process in refining and integrating these insights. Therefore, the setup presented should be seen solely as a tool for providing valuable leads for investigative data reporting, not as a replacement for the data reporting itself.

\bibliography{bibliography}


\begin{thebibliography}{15}


\ifx \showCODEN    \undefined \def \showCODEN     #1{\unskip}     \fi
\ifx \showDOI      \undefined \def \showDOI       #1{#1}\fi
\ifx \showISBNx    \undefined \def \showISBNx     #1{\unskip}     \fi
\ifx \showISBNxiii \undefined \def \showISBNxiii  #1{\unskip}     \fi
\ifx \showISSN     \undefined \def \showISSN      #1{\unskip}     \fi
\ifx \showLCCN     \undefined \def \showLCCN      #1{\unskip}     \fi
\ifx \shownote     \undefined \def \shownote      #1{#1}          \fi
\ifx \showarticletitle \undefined \def \showarticletitle #1{#1}   \fi
\ifx \showURL      \undefined \def \showURL       {\relax}        \fi
\providecommand\bibfield[2]{#2}
\providecommand\bibinfo[2]{#2}
\providecommand\natexlab[1]{#1}
\providecommand\showeprint[2][]{arXiv:#2}

\bibitem[Broussard(2015)]%
        {broussard2015artificial}
\bibfield{author}{\bibinfo{person}{Meredith Broussard}.} \bibinfo{year}{2015}\natexlab{}.
\newblock \showarticletitle{Artificial intelligence for investigative reporting: Using an expert system to enhance journalists’ ability to discover original public affairs stories}.
\newblock \bibinfo{journal}{\emph{Digital journalism}} \bibinfo{volume}{3}, \bibinfo{number}{6} (\bibinfo{year}{2015}), \bibinfo{pages}{814--831}.
\newblock


\bibitem[Cheng et~al\mbox{.}(2023)]%
        {cheng2023gpt}
\bibfield{author}{\bibinfo{person}{Liying Cheng}, \bibinfo{person}{Xingxuan Li}, {and} \bibinfo{person}{Lidong Bing}.} \bibinfo{year}{2023}\natexlab{}.
\newblock \showarticletitle{Is GPT-4 a Good Data Analyst?}. In \bibinfo{booktitle}{\emph{Findings of the Association for Computational Linguistics: EMNLP 2023}}. \bibinfo{pages}{9496--9514}.
\newblock


\bibitem[Diakopoulos(2020)]%
        {diakopoulos2020computational}
\bibfield{author}{\bibinfo{person}{Nicholas Diakopoulos}.} \bibinfo{year}{2020}\natexlab{}.
\newblock \showarticletitle{Computational news discovery: Towards design considerations for editorial orientation algorithms in journalism}.
\newblock \bibinfo{journal}{\emph{Digital journalism}} \bibinfo{volume}{8}, \bibinfo{number}{7} (\bibinfo{year}{2020}), \bibinfo{pages}{945--967}.
\newblock


\bibitem[Diakopoulos et~al\mbox{.}(2020)]%
        {diakopoulos2020generating}
\bibfield{author}{\bibinfo{person}{Nicholas Diakopoulos}, \bibinfo{person}{Madison Dong}, \bibinfo{person}{Leonard Bronner}, {and} \bibinfo{person}{Jeremy Bowers}.} \bibinfo{year}{2020}\natexlab{}.
\newblock \showarticletitle{Generating location-based news leads for national politics reporting}. In \bibinfo{booktitle}{\emph{Proc Computational+ Journalism Symposium}}.
\newblock


\bibitem[Diakopoulos et~al\mbox{.}(2021)]%
        {diakopoulos2021towards}
\bibfield{author}{\bibinfo{person}{Nicholas Diakopoulos}, \bibinfo{person}{Daniel Trielli}, {and} \bibinfo{person}{Grace Lee}.} \bibinfo{year}{2021}\natexlab{}.
\newblock \showarticletitle{Towards understanding and supporting journalistic practices using semi-automated news discovery tools}.
\newblock \bibinfo{journal}{\emph{Proc. of the ACM on Human-Computer Interaction}} \bibinfo{volume}{5}, \bibinfo{number}{CSCW2} (\bibinfo{year}{2021}), \bibinfo{pages}{1--30}.
\newblock


\bibitem[Hamilton and Turner(2009)]%
        {hamilton2009accountability}
\bibfield{author}{\bibinfo{person}{James~T Hamilton} {and} \bibinfo{person}{Fred Turner}.} \bibinfo{year}{2009}\natexlab{}.
\newblock \showarticletitle{Accountability through algorithm: Developing the field of computational journalism}. In \bibinfo{booktitle}{\emph{Report from the Center for Advanced Study in the Behavioral Sciences, Summer Workshop}}. \bibinfo{pages}{27--41}.
\newblock


\bibitem[Harcup and O'Neill(2016)]%
        {Harcup:2016bn}
\bibfield{author}{\bibinfo{person}{Tony Harcup} {and} \bibinfo{person}{Deirdre O'Neill}.} \bibinfo{year}{2016}\natexlab{}.
\newblock \showarticletitle{{What is news? News values revisited (again)}}.
\newblock \bibinfo{journal}{\emph{Journalism Studies}} \bibinfo{volume}{23}, \bibinfo{number}{1} (\bibinfo{date}{03} \bibinfo{year}{2016}), \bibinfo{pages}{1 -- 19}.
\newblock


\bibitem[Nishal et~al\mbox{.}(2024a)]%
        {nishal2024domain}
\bibfield{author}{\bibinfo{person}{Sachita Nishal}, \bibinfo{person}{Charlotte Li}, {and} \bibinfo{person}{Nicholas Diakopoulos}.} \bibinfo{year}{2024}\natexlab{a}.
\newblock \showarticletitle{Domain-Specific Evaluation Strategies for AI in Journalism}.
\newblock \bibinfo{journal}{\emph{arXiv preprint arXiv:2403.17911}} (\bibinfo{year}{2024}).
\newblock


\bibitem[Nishal et~al\mbox{.}(2024b)]%
        {nishal2024understanding}
\bibfield{author}{\bibinfo{person}{Sachita Nishal}, \bibinfo{person}{Jasmine Sinchai}, {and} \bibinfo{person}{Nicholas Diakopoulos}.} \bibinfo{year}{2024}\natexlab{b}.
\newblock \showarticletitle{Understanding Practices around Computational News Discovery Tools in the Domain of Science Journalism}.
\newblock \bibinfo{journal}{\emph{Proceedings of the ACM on Human-Computer Interaction}} \bibinfo{volume}{8}, \bibinfo{number}{CSCW1} (\bibinfo{year}{2024}), \bibinfo{pages}{1--36}.
\newblock


\bibitem[{OpenAI}(2024)]%
        {openai2024}
\bibfield{author}{\bibinfo{person}{{OpenAI}}.} \bibinfo{year}{2024}\natexlab{}.
\newblock \bibinfo{title}{OpenAI Assistants API v2 Documentation}.
\newblock
\newblock
\urldef\tempurl%
\url{https://platform.openai.com/docs/assistants/overview}
\showURL{%
\tempurl}
\newblock
\shownote{Accessed: 2024-05-14}.


\bibitem[Park et~al\mbox{.}(2023)]%
        {park2023generative}
\bibfield{author}{\bibinfo{person}{Joon~Sung Park}, \bibinfo{person}{Joseph O'Brien}, \bibinfo{person}{Carrie~Jun Cai}, \bibinfo{person}{Meredith~Ringel Morris}, \bibinfo{person}{Percy Liang}, {and} \bibinfo{person}{Michael~S Bernstein}.} \bibinfo{year}{2023}\natexlab{}.
\newblock \showarticletitle{Generative agents: Interactive simulacra of human behavior}. In \bibinfo{booktitle}{\emph{Proceedings of the 36th Annual ACM Symposium on User Interface Software and Technology}}. \bibinfo{pages}{1--22}.
\newblock


\bibitem[Rasheed et~al\mbox{.}(2024)]%
        {rasheed2024can}
\bibfield{author}{\bibinfo{person}{Zeeshan Rasheed}, \bibinfo{person}{Muhammad Waseem}, \bibinfo{person}{Aakash Ahmad}, \bibinfo{person}{Kai-Kristian Kemell}, \bibinfo{person}{Wang Xiaofeng}, \bibinfo{person}{Anh~Nguyen Duc}, {and} \bibinfo{person}{Pekka Abrahamsson}.} \bibinfo{year}{2024}\natexlab{}.
\newblock \showarticletitle{Can Large Language Models Serve as Data Analysts? A Multi-Agent Assisted Approach for Qualitative Data Analysis}.
\newblock \bibinfo{journal}{\emph{arXiv preprint arXiv:2402.01386}} (\bibinfo{year}{2024}).
\newblock


\bibitem[Showkat and Baumer(2021)]%
        {showakat_baumer2021}
\bibfield{author}{\bibinfo{person}{Dilruba Showkat} {and} \bibinfo{person}{Eric Baumer}.} \bibinfo{year}{2021}\natexlab{}.
\newblock \showarticletitle{{Where Do Stories Come From? Examining the Exploration Process in Investigative Data Journalism}}.
\newblock \bibinfo{journal}{\emph{Proc. ACM (CSCW)}} \bibinfo{volume}{5}, \bibinfo{number}{CSCW2} (\bibinfo{year}{2021}), \bibinfo{pages}{1--31}.
\newblock


\bibitem[Stigall et~al\mbox{.}(2023)]%
        {stigall2023chatbots}
\bibfield{author}{\bibinfo{person}{Brodrick Stigall}, \bibinfo{person}{Ryan Rossi}, \bibinfo{person}{Jane Hoffswell}, \bibinfo{person}{Xiang Chen}, \bibinfo{person}{Shunan Guo}, \bibinfo{person}{Fan Du}, \bibinfo{person}{Eunyee Koh}, {and} \bibinfo{person}{Kelly Caine}.} \bibinfo{year}{2023}\natexlab{}.
\newblock \showarticletitle{On Chatbots for Visual Exploratory Data Analysis}. In \bibinfo{booktitle}{\emph{2023 IEEE International Conference on Big Data (BigData)}}. IEEE, \bibinfo{pages}{5924--5929}.
\newblock


\bibitem[Wang et~al\mbox{.}(2019)]%
        {Wang2019}
\bibfield{author}{\bibinfo{person}{Yun Wang}, \bibinfo{person}{Zhida Sun}, \bibinfo{person}{Haidong Zhang}, \bibinfo{person}{Weiwei Cui}, \bibinfo{person}{Ke Xu}, \bibinfo{person}{Xiaojuan Ma}, {and} \bibinfo{person}{Dongmei Zhang}.} \bibinfo{year}{2019}\natexlab{}.
\newblock \showarticletitle{{DataShot: Automatic Generation of Fact Sheets from Tabular Data}}.
\newblock \bibinfo{journal}{\emph{Computer}} (\bibinfo{year}{2019}), \bibinfo{pages}{1 -- 1}.
\newblock


\end{thebibliography}
\end{document}